\documentclass[preprint,12pt]{elsarticle}

\usepackage{graphicx}
\usepackage{amssymb}

\usepackage{adjustbox}

\usepackage{epsfig}
\usepackage{amsmath}
\usepackage{multirow}

\journal{Journal Name}

\begin{document}

\begin{frontmatter}


\title{Spectrum Translation for Cross-Spectral Ocular Matching}



\author{Kevin Hernandez Diaz, Fernando Alonso-Fernandez, Josef Bigun}

\address{Halmstad University, Sweden}

\begin{abstract}

Cross-spectral verification remains a big issue in biometrics, especially for the ocular area due to differences in the reflected features in the images depending on the region and spectrum used.

In this paper, we investigate the use of Conditional Adversarial Networks for spectrum translation between near infra-red and visual light images for ocular biometrics. We analyze the transformation based on the overall visual quality of the transformed images and the accuracy drop of the identification system when trained with opposing data. 

We use the PolyU database and propose two different systems for biometric verification, the first one based on Siamese Networks trained with Softmax and Cross-Entropy loss, and the second one a Triplet Loss network. We achieved an EER of 1\% when using a Triplet Loss network trained for NIR and finding the Euclidean distance between the real NIR images and the fake ones translated from the visible spectrum. We also outperform previous results using baseline algorithms.

\end{abstract}


\end{frontmatter}


\section{Introduction}
\label{S:1 Introd}

Biometric recognition has gained attention in the last years due to an increment of interest in secure systems for border control and forensic purposes but also for private reasons like access control to offices and electronic devices like computers and smartphones. In this context, deep learning systems like convolutional neural networks (CNNs) have contributed significantly to their extension due to their state of the art performance in many different signal processing, computer vision, and pattern recognition tasks \citet{DLforbiometricsurvey}. 

Over the years, many different body regions have been proposed as potential biometric recognition cues like fingerprints, voice, hand geometry, gait, face, iris...\citet{jain2004introduction}. The ocular area has proven to be not only one of the most discriminative regions of the face, but it also gives quite stable performance among population and age. Many researchers have investigated the different parts of this area for biometric purposes \citet{surveyocularbio}. E.g. iris's texture pattern recognition \citet{howiriswork}, vein patterns in the sclera \citet{sclerarecogn} or retina \citet{retinapequeno} and the general shape and texture of the periocular area \citet{park2009periocular}. Other newer and more complex cues based on ocular movements have arisen as biometric options due to their distinctive patterns and their difficulties to spoof. Some examples on what this type of algorithms look are tracking eye movements or gaze due to certain stimulus \citet{eyemovement}, eye muscle movements \citet{eyemusclemovement}, or eye fixation and saccades \citet{eyesaccades}.

Depending on which ocular biometric you are using and the acquisition constraint of the system, a specific spectral bandwidth can achieve better performance. In the case of iris recognition, the use of Near-Infrared (NIR) cameras allows extracting more detailed textures images of the iris pattern, especially when dealing with dark irises \citet{howiriswork}. However, RGB images of iris can still be used for recognition depending on the circumstances, e.g., when the acquisition distance has to be more significant. Other systems like periocular or sclera are usually collected under visual light conditions (VIS) in RGB or grayscale images. For the sclera, standard cameras can capture the vein patterns in the upper part of this region, and more specifically, the use of the blue and green channels gives better performance than NIR cameras used for Iris recognition \citet{ocularvisiblespectrum}. The periocular area has been studied under NIR and VIS light \citet{alonso2016survey}. Under visible light, one can use further acquisition distances and detect better skin texture; however, NIR can also achieve good performance and be easily combined with the iris in multimodal biometric systems \citet{alonsoperiocularirisfusion}. 

In this paper, we focus on the periocular area and iris pattern for cross-spectral biometric recognition. The usefulness of such systems relies upon their flexibility, allowing good performance in long or short distances, change on the type of camera used, security requirements, or when dealing with legacy images from previous or different systems. More concretely, we analyze the use of Conditional Generative Adversarial Networks (CGANs) \citet{mirza2014conditional} for Image-to-Image transformations, in this case change of spectrum, in terms of visual and recognition quality. We also investigate the opportunities that this brings for biometric verification algorithms since it would allow closing the gap between the difference in the spectrum images that makes cross-spectral verification systems to perform worse than same-spectrum matching.

The rest of the paper is organized as follows: Section \ref{S:2 Related} summarizes related research work in the field of cross-spectral ocular recognition; in Section \ref{S:3 exp} we explain the different parts involved in the experimentation of this research; in Section \ref{S4: Res} we present the results from all of our experiments; Section \ref{S5: Diss} analyze the results of the experiments, try to extract any insight from them and propose future work based on them.

\section{Related Work}
\label{S:2 Related}

In this section, we will present some previous work related to this paper, with a focus on neural networks and cross-spectral biometric recognition.

In comparison to face and iris, cross-spectral periocular recognition research is relatively new. In \citet{oncrosspsectral}, the authors provide a new dataset for cross-spectral periocular recognition, consisting of images captured under the visible light spectrum, near-infrared, and night vision. They preprocessed the images to extract the pyramid of histogram of oriented gradients (PHOG) \citet{PHOG}, trained a NN for each spectrum, and then combined the inputs and the networks for cross-spectral recognition. In 2016, Cross-eyed \citet{crosseyed2016}, a competition on cross-spectral periocular and iris recognition, called researchers to show recent progress in the field. They provided two different datasets one for periocular and other for iris, both of them captured with a dual spectrum sensor. The best results obtained for each biometric treat was of an EER of 0.29\% for periocular and 2.78\% for iris, showing the difficulties of comparing iris images captured in different spectra due to iris melanin, in contrast to the rather well performance of the periocular area.

In \citet{nalla2016toward}, the creators of the PolyU dataset used in this paper, proposed two different frameworks for cross-spectral iris recognition. The first one based on Domain Adaptation using Naïve-Bayesian Nearest Neighbor (DA-NBNN) and the other based on Markov random field models (MRF). Later on, in \citet{polyuhashing} they trained a CNN trained with softmax and cross-entropy loss for feature extraction and two approaches for classification, joint-Bayesian inference, and supervised discrete hashing algorithm for dimension reduction and Hamming distance between vectors for classification.

In \citet{depresion}, they use pretrained models first trained for the image-net dataset, with a later fine-tuning for face recognition with the MS-Celeb-1M \citet{celeb}, a database consisting on 10M images, and afterward another fine-tuning step with the VGGFace2 dataset \citet{cao2018vggface2}, another face recognition database with a total of 3.31M images. Finally, they train the CNNs with the PolyU database for biometric identification using images of both spectra and then extracting a deep representation of the data by feeding the images to the network, extracting the semi-last layer and use cosine similitude to run biometric verification. This two steps training process and deep representation comparison for biometric recognition applications have achieved state of the art performance for other biometric traits as proposed in \citet{vggface2parairis} for iris recognition. 

In \citet{koch2015siamese}, the authors develop a siamese neural network structured based on weight sharing for one-shot image recognition. This architecture and some of its variations have been successfully used for biometric verification purposes, e.g., for face recognition in \citet{appliedsiamese}. Siamese networks have led the way to Triplet loss networks \citet{triplet-loss-networks}, being both distance-based classification networks. In \citet{tripletfacecross}, the authors use a triplet loss network with hard triplet selection based on similarity scores and transfer learning for cross-spectral face verification.

Some researchers have tried to create generative models to transform images from one spectrum to another. In \citet{thermal2visibleSGGAN}, they proposed a new model called Semantic-Guided GAN to transform from thermal to visible that uses semantic cues from the visible images and include them in the loss function for training. In \citet{cycleganthermalface} they analyze the use of GANs for spectrum translation between thermal and visible images for face recognition using Cycle-GANs and pix2pix and try to improve performance by adding loss term from the detection of landmark points in the faces.

\section{Experimentation}
\label{S:3 exp}
In this section, we describe the experimentation done in the paper. We divide it into the following subsections: Database, where we present the data used and any preprocessing done; Spectrum Translation where we explain the transformation algorithm and Biometric Verification, where we state the different matching algorithms tried.

\subsection{Database}

\begin{figure}[h!]
    \centering
    \includegraphics[width=\textwidth]{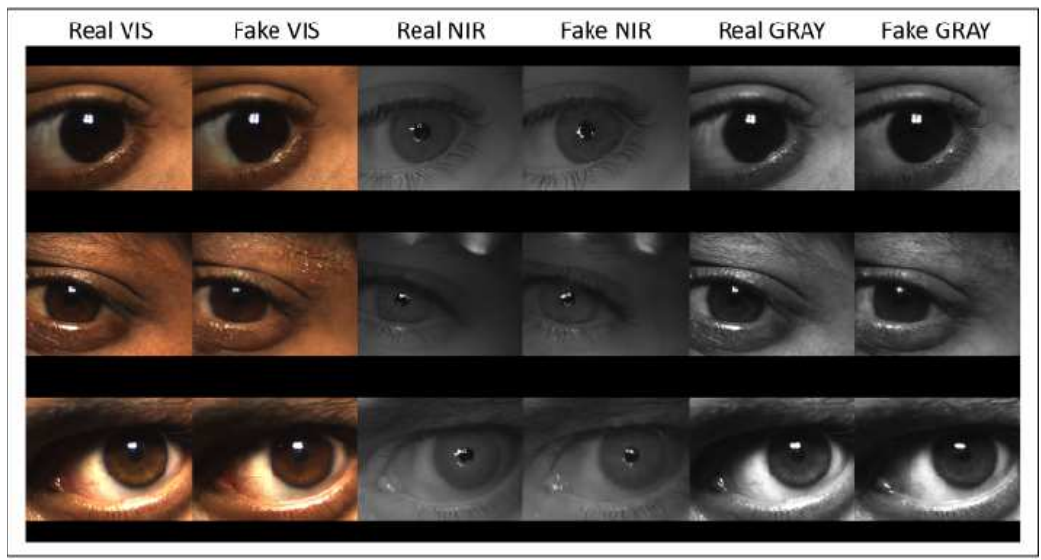}
    \caption{Example of periocular images in the database and their corresponding transformation after the spectrum translation}
    \label{fig:BBDD_perioc}
\end{figure}

\begin{figure}[h!]
    \centering
    \includegraphics[width=\textwidth]{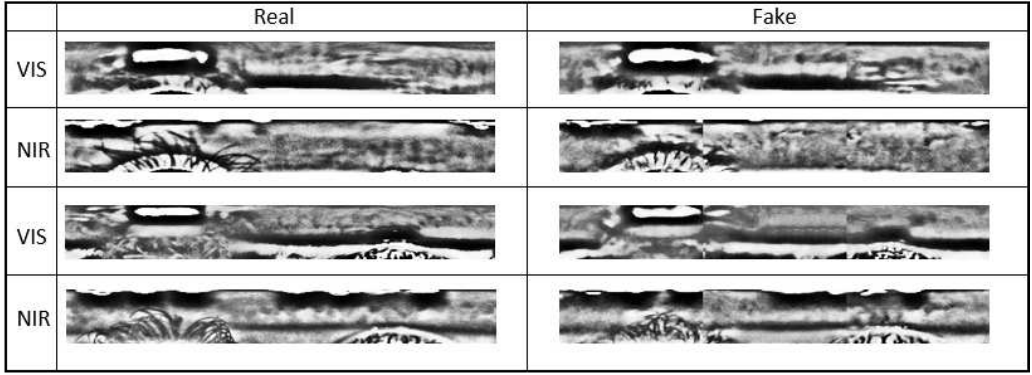}
    \caption{Example of Iris images in the database and their corresponding transformation after the spectrum translation. The first two rows belongs to one ID and the last two to another oen}
    \label{fig:BBDD_iris}
\end{figure}

For this work, we made use of "The Hong Kong Polytechnic University Cross-Spectral Iris Images Database" \citet{nalla2016toward} with the purpose of cross-spectral biometric verification. The database consists of 15 images of both iris identities from 209 subjects caught in two different spectra (Near Infra-Red and Visible Light), making a total of 12,540 pictures (15 images x 2 spectrum x 2 iris x subject). Each one has a resolution of 640x480 pixels and has perfect pixel correspondence between the two spectra. It also provides the normalized irises in images of 64x512 pixels as well as an enhanced version. We divided the dataset using the first 10 images of each eye for training and the last 5 for testing, as done in the original database's paper \citet{nalla2016toward}. An example of the data can be seen in Figure \ref{fig:BBDD_perioc} in the real columns for the periocular images and in Figure \ref{fig:BBDD_iris} for the enhanced irises. Each row in Figure \ref{fig:BBDD_perioc} belongs to one subject ID, and in Figure \ref{fig:BBDD_iris} the first two rows belong to one subject and the last two to another.

For the periocular region, we made the images squared by adding a padding region fill with zeroes to avoid deformation of the images and then downsampled them to 256x256 pixels. We did this to fulfill the network requirements of the spectrum translation algorithms that we will explain in the corresponding section. No further preprocessing steps were taken due to such small and irregular periocular region in the images.

\begin{figure}[h!]
    \centering
    \includegraphics[width=\textwidth]{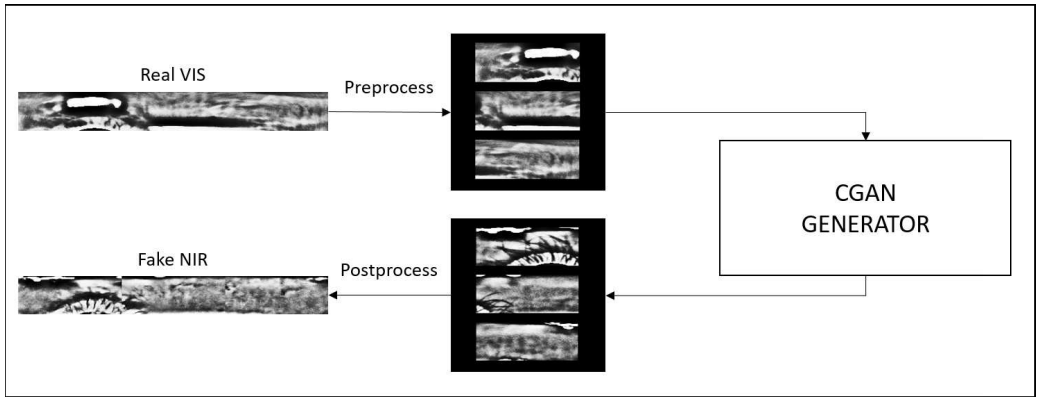}
    \caption{Example of iris preprocessing}
    \label{fig:Iris spectrum}
\end{figure}

For the iris experiments, we used the enhanced iris version given with the dataset for results comparisons. To make the normalized images the correct size for the spectrum translation protocol, and to avoid such big deformation and information loss, we create an auxiliary middle image representation by creating a 256x256 image filled with zeroes, divide the iris image into 3 parts of ~64x171 pixels and fit it in. 
A similar approach to this method can be seen in \citet{paper_iris_concatenated}, where they divide the iris into upper and lower parts and concatenate the normalized images on top of each other. After the image has been transformed into the other spectrum, we return the image to the original size by extracting the regions and putting them back together. All this process can be seen in Figure \ref{fig:Iris spectrum}.

\subsection{Spectrum Translation}

In this section, we describe the spectrum translation algorithm used in the experiments. The goal of the spectrum translation is to close the gap between images captured in different spectrum representations to make it look as if they were taken in the other spectrum. To do this, we make use of Conditional Generative Adversarial Networks (CGANs) \citet{mirza2014conditional}. Contrarily to traditional GANs, that learn to map an image from a random noise matrix, CGANs are conditioned to some extra information in the input. In our case, the network input is either a periocular or iris image captured in a specific spectrum. The loss function of a CGAN is commonly expressed as in equation \ref{objectivecGAN}, where the generator tries to minimize the recognition rate of the discriminator by creating more realistic images. The discriminator, on the other hand, tries to maximize its recognition rate of fake generated images.

\begin{figure}[h!]
    \centering
    \includegraphics[width=\textwidth]{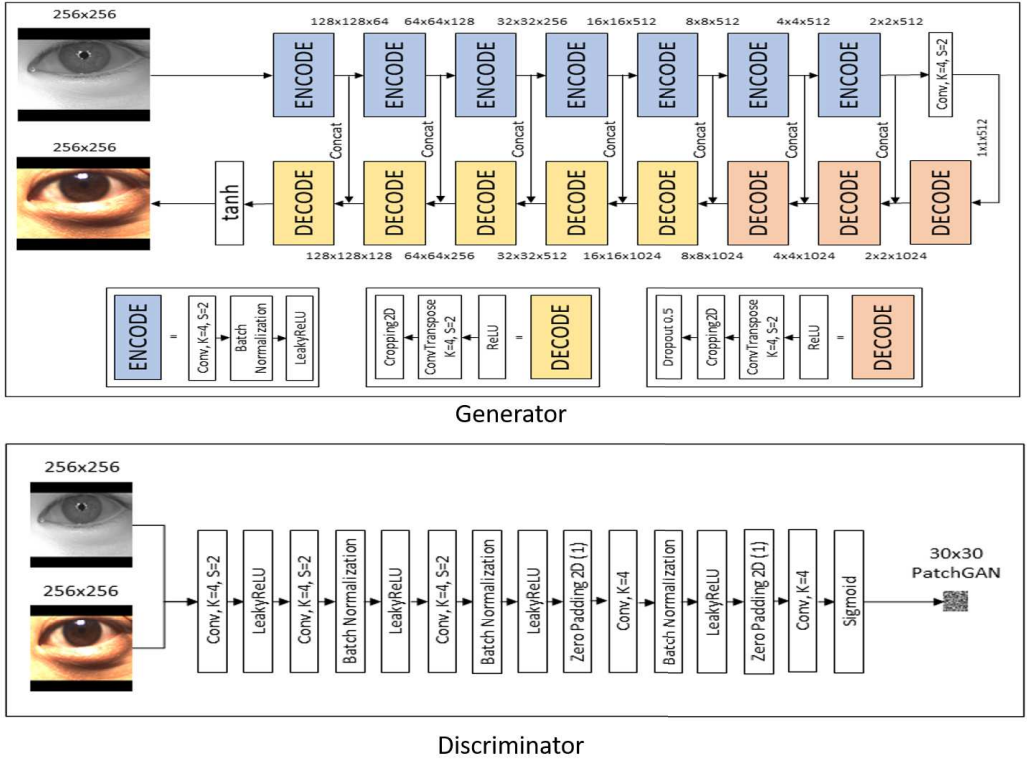}
    \caption{Pix2Pix architectures}
    \label{Pix2Pixarchitectures}
\end{figure}

\begin{equation}
\label{objectivecGAN}
\begin{aligned}
    \mathcal{L}_{cGAN}(G,D)= & \mathbb{E}_{x,y}[log D(x,y)] + \\ & \mathbb{E}_{x,y}[log (1 - D(x,G(x,z))]
\end{aligned}
\end{equation}

We make use of one of the most famous CGAN for image-to-image translation called Pix2Pix \citet{isola2017image}. In contrast to other solutions like \citet{cyclegan} where the network learns a mapping based on the general look of the inputs and output images, this network force the generator's mapping to be close to the target image by adding to the loss function a L1 term as expressed in equation \ref{L1norm}, leaving the discriminator's loss unchanged. Another reason for this choice is that since the data have perfect pixel alignment, it should ease the work of the generator to get a closer representation of the output and help to minimize the L1 loss in equations \ref{L1norm}. Giving a more accurate representation of the target data. The final loss function of the CGAN can be seen in equation \ref{loss}

\begin{equation}
\label{L1norm}
    \mathcal{L}_{L1}(G)=\mathbb{E}_{x,y,z}[\|y-G(x,z)\|_1]
\end{equation}

\begin{equation}
\label{loss}
    G^{*}=arg \min_{G}\max_{D}\mathcal{L}_{cGAN}(G,D)+\lambda\mathcal{L}_{L1}(G)
\end{equation}

For this project, we use a Keras implementation of Pix2Pix \footnote{https://github.com/tjwei/GANotebooks}. The generator uses a version of the famous encoder-decoder architecture U-NET \citet{ronneberger2015u}, in this particular case, a U-NET-256 with batch normalization. The discriminator is a PatchGAN, where the output is an image of size 30x30. Each pixel in the output image corresponds to a patch of size 70x70 in the input. This 70x70 PatchGAN has proven to give sharper results both in spatial and spectral dimensions. Finally, the discriminator calculates the loss by checking that each patch is correctly classified or not and updates its parameters using Adam as the optimizer. The generator and discriminator's architectures can be seen in Figure \ref{Pix2Pixarchitectures}

\subsection{Biometric Verification}

In this section, we describe the main biometric recognition experiments. We follow the same procedure for both periocular and iris experiments. After the spectrum translation is done, we trained a ResNet50 \citet{he2016deep}, pretrained with the ImageNet\footnote{http://www.image-net.org/} dataset to run biometric identification. We use the first 10 images of every user for training and the remaining 5 for testing. There are three reasons for this, firstly, to asses the goodness of the spectrum translation algorithm through the accuracy drop when trained with real data and test it with both real and fake images. Secondly, this will allow us to reduce the amount of time and computational power when training the triplet loss network for the biometric verification experiments. Finally, we believe that this will make the triplet loss network to be less prone to find image differences as alignment and scale instead of biometric features. The importance of this is due to the high inter-class variation and low intra-class variation present in the DDBB, and no room for alignment/scale preprocessing steps since the periocular area is so small and irregular. Figure \ref{fig:BBDD_perioc} show an example of this phenomenon.

We propose 2 different biometric verification models. The first one consists of a double-headed CNN architecture, each branch being a previously trained CNN for biometric identification with a specific type of data input, as stated before. We froze the layers and concatenated the final embedded vector, then trained a fully connected neural network to classify the data as a genuine user or impostor. We used soft-max as the last activation function and trained the system with binary cross-entropy loss and Adam as the optimizer. 

\begin{figure}[h!]
    \centering
    \includegraphics[width=\textwidth]{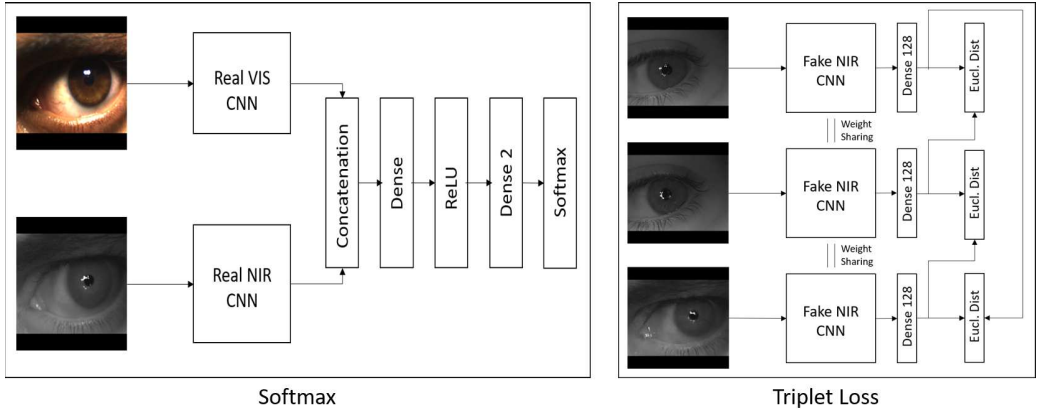}
    \caption{architectures examples}
    \label{architectures}
\end{figure}

The second system is a triplet loss network \citet{triplet-loss-networks}. Triplet loss networks have proven to give state of the art performance in many different tasks, and they are specially designed for image comparisons, which makes them a perfect fit for biometric verification systems. We make use of our previously trained CNNs for biometric identification as the base with their layers frozen. The main difference between the first proposal is that triplet loss networks need a similar type of data as inputs since the CNN architecture share the same weights along all branches. So for every triplet loss network tried, we create the inputs using a real image in one spectrum, a translated image of the same user from the other spectrum, and another fake image from an impostor, an example of an input image is shown in Figure \ref{architectures}. Finally, we train a fully connected network to create an optimized small vector representation for this task with its weights shared with every branch. We use Euclidean distance and the improved triplet loss function \citet{improved-triplet-loss} and Adam as the optimizer. The main difference between the improved triplet loss function and the traditional one is that it also maximizes the distance between the negative image and the positive one as expressed in equation \ref{improved-triplet-loss-eq}

\begin{equation}
\label{improved-triplet-loss-eq}
    L=max(0, D(x_{a},x_{p}) -  \frac{(D(x_{a},x_{n}) + D(x_{p},x_{n}))}{2} + \alpha)
\end{equation}

\section{Results}
\label{S4: Res}

In this section, we will briefly present the results found in the experimentation process. 

In Figure \ref{fig:BBDD_perioc} we present examples of the original periocular data postprocessed as well as the fake data generated by pix2pix for 3 users ID in each spectrum.  Figure \ref{fig:BBDD_iris} and \ref{fig:Iris spectrum} show examples of the DDBB original iris data, post-processed images as explained in section \ref{S:3 exp}, and fake images generated by the spectrum translation algorithm. The preprocessing steps made by us are fairly simple and are done to prepare the data for the network's requirements for the Image-to-Image translation algorithms.

The quality assessment for the spectrum translation algorithm can be seen in Table \ref{table:PSNR} for visual quality and in Table \ref{table:identification} for the biometric quality. We assess the spectrum translation algorithm in terms of Peak Signal to Noise Ratio (PSNR) and Structural Similarity Index (SSIM) by the mean value and standard deviation of the test set. The biometric quality is analyzed by the accuracy difference of the identification system when tested with real and fake data, i.e. the accuracy drop of when the CNN was trained with real NIR data and then tested with both real NIR data and translated images.

\begin{table*}[h!]
\centering
\begin{adjustbox}{max width=\textwidth}
\begin{tabular}{|c|c||c|c||c|c||} 
 \cline{2-6}
 \multicolumn{1}{c|}{} & \multirow{2}{*}{Metric} & \multicolumn{2}{|c||}{Left} & \multicolumn{2}{|c|}{Right} \\ \cline{3-6}
 \multicolumn{1}{c|}{} & & PSNR & SSIM & PSNR & SSIM \\\hline\hline
 
 \multicolumn{1}{|c|}{\multirow{3}{*}{\rotatebox[origin=c]{90}{Perioc}}} & VIS2NIR & \textbf{31.1$\pm2.7$} & 0.913$\pm$0.034& \textbf{31.6$\pm$2.9}&0.920$\pm$0.036 \\\cline{2-6}
  & NIR2VIS & 27.2$\pm$2.9& \textbf{0.94$\pm$0.026}& 27.6$\pm$2.9&\textbf{0.942$\pm$0.024}  \\\cline{2-6}
 \multicolumn{1}{|c|}{}& NIR2GRAY & 27.6$\pm$3.0 &0.869$\pm$0.054 &28.1$\pm$3.0 &0.883$\pm$0.047 \\\hline\hline
 
  \multicolumn{1}{|c|}{\multirow{2}{*}{\rotatebox[origin=c]{90}{Iris}}} & VIS2NIR &11.3$\pm$1.3 & 0.233$\pm$0.082& 11.3$\pm$1.3&0.233$\pm$0.084 \\\cline{2-6}
  & NIR2VIS & \textbf{11.9$\pm$1.4}&\textbf{0.260$\pm$0.088} &11.9$\pm$1.5 &0.248$\pm$0.094  \\\hline
 
\end{tabular}
\end{adjustbox}
\caption{Visual quality assessment}
\label{table:PSNR}
\end{table*}

\begin{table}[h!]
\centering
\begin{adjustbox}{max width=\textwidth}
\begin{tabular}{| c|c|c | c | c| c|c|c|c|c| } 
 \cline{2-10}
 \multicolumn{1}{c|}{} & \multirow{2}{*}{Train data} & \multicolumn{4}{|c|}{Left} & \multicolumn{4}{|c|}{Right}\\
 \cline{3-10}
  \multicolumn{1}{c|}{}&  & Test Real & Test Fake & Difference & mean & Test Real & Test Fake & Difference & mean  \\  
 \hline
 \multirow{6}{*}{\rotatebox[origin=c]{90}{Periocular}} & real VIS & 99.7 & 94.8 & -4.9 & 97.3& 99.3 & 92.1& -7.2&95.7 \\ \cline{2-10}
 & fake VIS & 98.6 & 97.3 &+1.3 &98.0 & 98.8 & 96.9& +1.9 &97.9\\ \cline{2-10}
 & real NIR & 99.7 & 94.1 & -5.6& 96.9& 99.6 & 79.5&  -20.1&89.6\\ \cline{2-10}
 & fake NIR & 99.2 & 99.4 & -0.2& 99.3& 98.5 & 91.9&  +6.6&95.2\\ \cline{2-10}
 & real GRAY & 100 & 95.7   & +4.3& 97.9& 99.3 & 96.1&  -6.6&97.7\\ \cline{2-10}
 & fake GRAY & 99.1 & 97.7& +1.4& 98.4& 98.3 & 95.7&  +2.6&97\\ 
 \hline\hline
 \multirow{4}{*}{\rotatebox[origin=c]{90}{Iris}} & real VIS & 90.7 & 25.2 & -65.5& 58.0& 88.6 & 26.0& -62.6 &57.3\\ \cline{2-10}
 & fake VIS & 39.5 & 76.7 &-37.2 & 58.1& 39.7 & 76.8 &-37.1&58.3\\\cline{2-10}
 & real NIR & 93.5 & 40.1 & -53.4& 66.8& 91.2 & 45.7 &-45.5&68.5\\\cline{2-10}
 & fake NIR & 57.4 & 79.0   & -21.6& 68.2& 60.0 & 77.7 & -17.7&68.9\\
 \hline
\end{tabular}
\end{adjustbox}
\caption{biometric recognition quality assessment}
\label{table:identification}
\end{table}

For each biometric treat in the verification experiments, we first calculate baseline results using widely used algorithms. For these experiments, we combine the images, consider each eye as a different user and use the same partitions used for training and test, as explained in section \ref{S:2 Related}. For genuine scores, we make all against all comparisons with the test set, getting a total of 209 users * 2 eyes * 10 combinations = 4180 genuine scores, using the same combinations that we would make in same-spectrum recognition for cross-spectral comparisons. For impostor scores, we compare the first image of each user against the second image of the rest of the users, getting a total of 418 * 417 = 174306 impostor comparisons.  

We calculate baseline results used for biometric recognition for both periocular and iris as reference. We show the results obtained for same spectrum recognition in Table \ref{table:baselineSameSpectrum} and for cross-spectral in Table \ref{table:baselineCrossSpectrum}. 

For the periocular images we made use of commonly used algorithms for periocular recognition systems \citet{alonso2016survey} like Local Binary Pattern (LBP) \citet{LBP}, Histogram of Oriented Gradients (HOG) \citet{HOG} and Scale Invariant Feature Transform (SIFT) \citet{SIFT} as control results for our experiments. The authentication process for LBP and HOG consist of a simple distance measure between the images, while for SIFT we use the number of paired key-points normalized by the mean number of points in both images. We finally combine the score of the previous systems using a Linear Logistic Regression function, as explained in \citet{logisticregression}, to try to improve the results. 

For the case of iris recognition, we try 7 different algorithms that comes in the USIT toolkit v2 \citet{USIT2}. Concretely, we use Context-based (CB) algorithm \citet{rathgeb2011contextbase}, Complex Gabor filterbanks (CG) as used by Daugman \citet{daugmanCG}, Local Intesity Variations (CR) algorithm by \citet{rathgebCR}, Difference of Cosine Transform (DCT) coefficients developed by \citet{monro2007dct}, cumulative-sum-based change analysis for iris recognition systems developed by \citet{ko2007novel} (KO), 1D-Log Gabor wavelets (LG-1D) and Gabor Spatial Filters (QSW) \citet{qswIrisMa}. Context-based, Difference of Cosine Transform coefficients and the cumulative-sum-based change analysis use their specific comparison algorithms, while the rest use Hamming-Distance.

\begin{table*}[h!]
\centering
\label{table:baselineSameSpectrum}
\begin{adjustbox}{max width=\textwidth}
\begin{tabular}{|c|c||c|c|c||c|c|c|}

\cline{2-8}
\multicolumn{1}{c|}{} & \multirow{3}{*}{System} & \multicolumn{3}{|c|}{NIR} & \multicolumn{3}{|c|}{VIS} \\\cline{3-8}
\multicolumn{1}{c|}{} & & \multirow{2}{*}{EER} & GAR @ & GAR @  & \multirow{2}{*}{EER} & GAR @  & GAR @  \\
\multicolumn{1}{c|}{} & & & 1\% FAR & 0.1\% FAR& &  1\% FAR & 0.1\% FAR \\\hline

\multirow{6}{*}{\rotatebox[origin=c]{90}{Periocular}}& LBP eucl &  5.3 & 90.1&83.8 &  6.2 & 87.2&79.1\\\cline{2-8}
& LBP $X^{2}$ &   5.3 & 89.7&83.4&  5.8 & 89.0&81.4 \\\cline{2-8}
& HOG eucl  &   8.0 & 84.6&77.3&  9.0 & 82.4&75.0 \\\cline{2-8}
& HOG $X^{2}$ &   8.0 & 84.4&77.3&  9.3 & 82.2&74.6 \\\cline{2-8}
& SIFT      &   1.1 & 98.9&98.4&  1.6 & 98.3&97.3 \\\cline{2-8}
& Fusion eucl + SIFT &   1.1 & 98.9&98.5&  1.4 & 98.3&97.5 \\\cline{2-8}
& Fusion $X^{2}$ + SIFT &  1.1 & 98.9&98.5&  1.5 & 98.4&97.5 \\\hline

\multirow{6}{*}{\rotatebox[origin=c]{90}{Iris}}& CB \citet{rathgeb2011contextbase} &  23.7 & 46.1&7.5 &  32.0 & 9.2&0.0\\\cline{2-8}
& CG &   16.1 & 74.5&70.1&21.8 &59.1&49.3 \\\cline{2-8}
& CR  &   22.1&63.9&54.8&28.0&47.6&36.1 \\\cline{2-8}
& DCT &   16.1&65.3&50.0&24.4&44.7&24.9 \\\cline{2-8}
& KO     &   20.8&64.4&56.4&25.3&52.4&42.0 \\\cline{2-8}
& 1D-Log Gabor wavelets &   16.6&75.7&72.0&21.8&61.6&49.1 \\\cline{2-8}
& QSW &  16.6&76.0&72.3&23.2&61.1&53.7 \\\hline


\end{tabular}
\end{adjustbox}
\caption{Baseline results for Same Spectrum. All the results for iris were calculated using the systems and comparators availables in \citet{USIT2}. In order are: Context-based (CB) algorithm \citet{rathgeb2011contextbase}, Complex Gabor filterbanks (CG) as used by Daugman \citet{daugmanCG}, Local Intesity Variations (CR) algorithm \citet{rathgebCR}, Difference of Cosine Transform (DCT) coefficients \citet{monro2007dct}, cumulative-sum-based change analysis \citet{ko2007novel} (KO), 1D-Log Gabor wavelets (LG-1D) and Gabor Spatial Filters (QSW) \citet{qswIrisMa} }
\end{table*}

\begin{table*}[h!]
\centering
\label{table:baselineCrossSpectrum}
\begin{adjustbox}{max width=\textwidth}
\begin{tabular}{|c|c||c|c|c||c|c|c||c|c|c|}

\cline{2-11}
\multicolumn{1}{c|}{} & \multirow{3}{*}{System} & \multicolumn{3}{|c|}{real NIR \& fake NIR} & \multicolumn{3}{|c|}{real VIS \& fake VIS} & \multicolumn{3}{|c|}{real VIS \& real real NIR}\\\cline{3-11}
\multicolumn{1}{c|}{} & & \multirow{2}{*}{EER} & GAR @ & GAR @  & \multirow{2}{*}{EER} & GAR @  & GAR @  & \multirow{2}{*}{EER} & GAR @  & GAR @\\
\multicolumn{1}{c|}{} & & & 1\% FAR & 0.1\% FAR& &  1\% FAR & 0.1\% FAR & &  1\% FAR & 0.1\% FAR\\\hline

\multirow{8}{*}{\rotatebox[origin=c]{90}{Periocular}}& LBP eucl &  9.4 & 74.5&53.7&7.4&82.6&69.6&39.0&4.9&0.3 \\\cline{2-11}
& LBP $X^{2}$ &   12.2&63.2&38.8&7.0&83.3&68.6&39.0&4.7&0.7 \\\cline{2-11}
& HOG eucl  &   9.1&81.1&68.3&9.4&79.8&69.6&17.8&43.9&21.7 \\\cline{2-11}
& HOG $X^{2}$ &  9.4&80.2&66.8&9.7&79.3&69.0&19.6&37.0&16.4 \\\cline{2-11}
& SIFT      &   5.8&86.2&67.8&12.7&60.7&36.0&45.1&1.9&0.3 \\\cline{2-11}
& Fusion  Eucl.&    3.61 &  93.5 &  84.6 &  5.2 &  87.5 &  71.0 &  17.5 &  43.8 &  20.4 \\\cline{2-11}
& Fusion $X^{2}$ &    4.0 &  92.5 &  81.4 &  5.7 &  84.6 &  64.2 &  19.7 &  36.6 &  14.1 \\\hline\hline

\multirow{7}{*}{\rotatebox[origin=c]{90}{Iris}} & CB &    29.9 &  11.2 &  0.1 &  36.2 &  1.5 &  0.0 &  40.9 &  2.0 &  0.1 \\\cline{2-11}

& CG &    33.6 &  18.0 &  8.1 &  33.0 &  19.4 &  8.7 &  35.8 &  13.6 &  4.8 \\\cline{2-11}

& CR &    24.1 &  46.0 &  27.7 &  31.0 &  25.2 &  11.1 &  37.2 &  9.1 &  3.6 \\\cline{2-11}

& DCT &    37.6 &  12.4 &  4.0 &  39.0 &  11.4 &  1.1 &  41.7 &  3.1 &  0.4 \\\cline{2-11}

& KO &    24.6 &  43.4 &  25.8 &  27.5 &  33.6 &  15.6 &  38.3 &  7.1 &  1.8 \\\cline{2-11}

& 1D-Log &    26.1 &  43.5 &  30.8 &  28.4 &  33.2 &  16.7 &  31.0 &  28.8 &  15.5 \\\cline{2-11}

& QSW &    28.4 &  33.8 &  22.9 &  29.0 &  29.7 &  14.3 &  30.8 &  32.9 &  21.1 \\\hline


\end{tabular}
\end{adjustbox}
\caption{Baseline results for Cross Spectrum. All the results for iris were calculated using the systems and comparators availables in \citet{USIT2}. In order are: Context-based (CB) algorithm \citet{rathgeb2011contextbase}, Complex Gabor filterbanks (CG) as used by Daugman \citet{daugmanCG}, Local Intesity Variations (CR) algorithm \citet{rathgebCR}, Difference of Cosine Transform (DCT) coefficients \citet{monro2007dct}, cumulative-sum-based change analysis \citet{ko2007novel} (KO), 1D-Log Gabor wavelets (LG-1D) and Gabor Spatial Filters (QSW) \citet{qswIrisMa}}
\end{table*}

We present the results for all our experiments with CNNs in Table \ref{table:verification periocular} for the periocular images and in Table \ref{table:verification Iris} for the iris images. In the Softmax CNN, we extract the output of the last activation function and use the genuine score to calculate the metrics in the tables, while the Triplet Loss CNNs use euclidean distance between the final vector representation of each branch for authentification. For these experiments, we calculate the performance for the left and right eyes separately. We compare all images in one spectrum against all the images of the same user in the other spectrum to generate the genuine scores, getting a total of 209 users * 5 images in spectrum 1 * 5 images in spectrum 2 = 5225 genuine scores. For the impostor scores, we compare the first 5 images of a user against the first 5 images of the rest, respectively, getting a total of 209 users * 208 impostors * 5 images = 217360 impostor scores.

\begin{table*}[h!]
\centering
\label{table:verification periocular}
\begin{adjustbox}{max width=\textwidth}
\begin{tabular}{|c|c||c|c|c|c|c|c|c|c|}

\hline

\multirow{3}{*}{System} & \multirow{3}{*}{Data} & \multicolumn{4}{c|}{Left Eye} & \multicolumn{4}{c|}{Right Eye} \\ \cline{3-10}
& & \multirow{2}{*}{EER} & \multirow{2}{*}{\shortstack[l]{GAR @ \\ 10\% FAR}} & \multirow{2}{*}{\shortstack[l]{GAR @ \\ 1\% FAR}} & \multirow{2}{*}{\shortstack[l]{GAR @ \\ 0.1\% FAR}} & \multirow{2}{*}{EER} & \multirow{2}{*}{\shortstack[l]{GAR @ \\ 10\% FAR}} & \multirow{2}{*}{\shortstack[l]{GAR @ \\ 1\% FAR}} & \multirow{2}{*}{\shortstack[l]{GAR @ \\ 0.1\% FAR}} \\

& & & & & & & & & \\

\hline\hline

\multirow{5}{*}{Softmax} & VIS - NIR & 1.7 & 99.8 & 97.2 & 72.4 & 1.3 & 1.0 & 98.1 & 85.7 \\ \cline{2-10}
& VIS - F VIS & 1.65 & 99.7 & 97.3 & 85.5 & 2.24 & 99.8 & 95.12 & 75.4 \\ \cline{2-10}
& NIR - F NIR & 2.1 & 99.8 & 94.5 & 64.0 & 0.9 & 1.0 & 99.2 & 89.5 \\ \cline{2-10}
& GRAY - F GRAY & 1.4 & 99.9 & 98.0 & 89.9 & 2.93 & 99.6 & 91.0 & 60.6 \\ \cline{2-10}
& GRAY - NIR & 1.8 & 99.9 & 96.2 & 63.3 & 2.26 & 99.9 & 94.6 & 58.0 \\ 
\hline\hline

\multirow{6}{*}{\shortstack[l]{Triplet \\ Loss}} & VIS (R) & 1.5 & 99.7 & 97.8 & 92.9 & 1.95 & 99.6 & 97.2 & 89.8 \\ \cline{2-10}
& VIS (F) & 2.3 & 99.4 & 96.5 & 92.3 & 1.3 & 99.6 & 98.4 & 93.6 \\ \cline{2-10}
& NIR (R) & 1.5 & 99.6 & 98.2 & 93.5 & 1.7 & 99.5 & 97.8 & 92.5 \\ \cline{2-10}
& NIR (F) & 0.9 & 99.8 & 99.2 & 97.1 & 1.0 & 99.8 & 99.0 & 96.6 \\ \cline{2-10}
& GRAY (R) & 1.4 & 99.9 & 98.3 & 95.9 & 1.7 & 99.6 & 97.5 & 94.0 \\ \cline{2-10}
& GRAY (F) & 1.3 & 99.7 & 98.5 & 95.9 & 2.2 & 99.6 & 96.8 & 90.4 \\ 

\hline

\hline
\end{tabular}
\end{adjustbox}
\caption{Periocular verification system}
\end{table*}

\begin{table*}[h!]
\centering
\label{table:verification Iris}
\begin{adjustbox}{max width=\textwidth}
\begin{tabular}{|c|c||c|c|c|c|c|c|c|c|}

\hline

\multirow{3}{*}{System} & \multirow{3}{*}{Data} & \multicolumn{4}{c|}{Left Eye} & \multicolumn{4}{c|}{Right Eye} \\ \cline{3-10}
& & \multirow{2}{*}{EER} & \multirow{2}{*}{\shortstack[l]{GAR @ \\ 10\% FAR}} & \multirow{2}{*}{\shortstack[l]{GAR @ \\ 1\% FAR}} & \multirow{2}{*}{\shortstack[l]{GAR @ \\ 0.1\% FAR}} & \multirow{2}{*}{EER} & \multirow{2}{*}{\shortstack[l]{GAR @ \\ 10\% FAR}} & \multirow{2}{*}{\shortstack[l]{GAR @ \\ 1\% FAR}} & \multirow{2}{*}{\shortstack[l]{GAR @ \\ 0.1\% FAR}} \\

& & & & & & & & & \\

\hline\hline

\multirow{3}{*}{Softmax} & VIS - NIR & 9.5 & 91.0 & 66.5 & 22.6 & 10.3 & 89.5 & 69.7 & 40.4 \\ \cline{2-10}
& VIS - F VIS & 11.0 & 88.1 & 61.1 & 26.4 & 11.9 & 86.7 & 62.3 & 27.5 \\ \cline{2-10}
& NIR - F NIR & 9.8 & 90.2 & 68.8 & 37.1 & 11.5 & 87.3 & 62.8 & 33.2 \\ 
\hline\hline

\multirow{4}{*}{\shortstack[l]{Triplet \\ Loss}} & VIS (R) & 18.7 & 73.2 & 42.6 & 18.6 & 21.8 & 67.4 & 37.8 & 17.1 \\ \cline{2-10}
& VIS (F) & 19.4 & 71.1 & 41.2 & 18.2 & 20.3 & 71.3 & 41.3 & 21.6 \\ \cline{2-10}
& NIR (R) & 15.6 & 80.1 & 55.3 & 31.2 & 16.7 & 79.5 & 61.0 & 40.4 \\ \cline{2-10}
& NIR (F) & 15.9 & 78.9 & 52.2 & 27.5 & 16.4 & 78.9 & 55.1 & 32.3 \\ 

\hline

\hline
\end{tabular}
\end{adjustbox}
\caption{Iris verification system}
\end{table*}

\begin{figure}[h!]
    \centering
    \includegraphics[width=\textwidth]{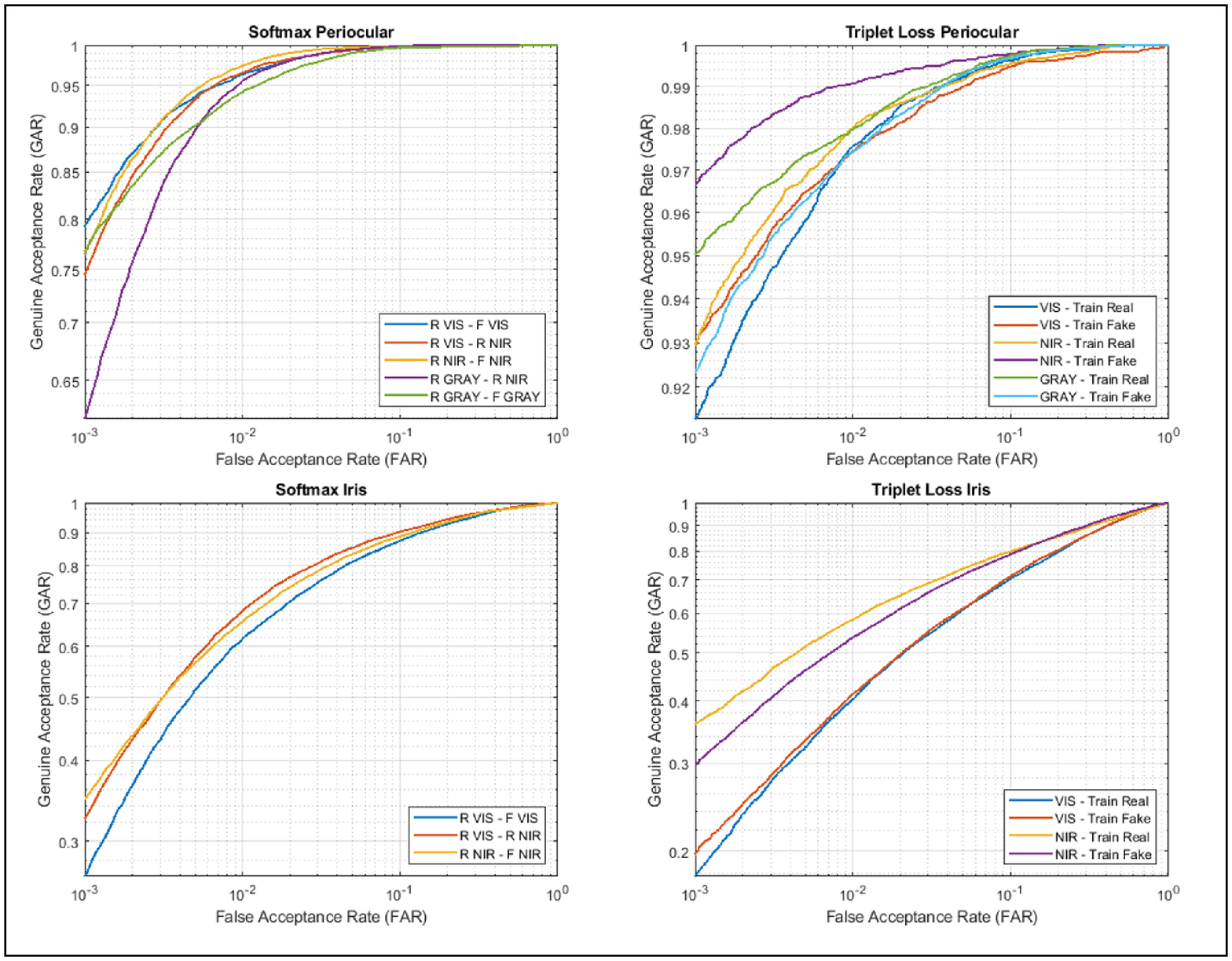}
    \caption{ROC curves}
    \label{DETs}
\end{figure}

Finally, we compare the best results from our proposed systems with previously reported ones for the PolyU database in Table \ref{table:summary}. To obtain the results in this table, we combine the scores from both eyes, we concatenate them and then recalculate the metrics based on this new vector. We did this to get a better comparison with the references and to get a better representation on how the overall performance of the best system for each approach. We plot the ROC curves for each CNN system proposed in Figure \ref{DETs}.

\begin{table*}[h!]
\centering
\begin{adjustbox}{max width=\textwidth}
\begin{tabular}{||c|c | c || c|c|c|c||} 
 \hline
 System & Biometry &  Data & EER & GAR @ 10\% & GAR @ 1\% & GAR @ 0.1\% \\  
 
 \hline\hline
 Fusion Baseline & Periocular &NIR & 3.6 & x & 93.5 & 84.6 \\\hline
 CR Baseline & Iris &NIR & 24.1 & x & 46.0 & 27.7 \\\hline
 Softmax & Periocular & NIR  & 1.5 & \textbf{99.9}& 97.4& 76.4\\\hline
 TL & Periocular & NIR (F)  &\textbf{1} & 99.8&\textbf{99.1} &\textbf{96.7} \\\hline
 Softmax & Iris &VIS\&NIR  & 9.8& 90.3& 68.1&31.1 \\\hline
 TL & Iris & NIR (R)  & 16.2&79.9 &58.3 &36.0 \\\hline\hline
 \citet{nalla2016toward}* & Iris & VIS\&NIR &  33.89 & 61.9 & $\sim$33.0 & $\sim$20.0 \\\hline
 \citet{polyuhashing}* & Iris & VIS\&NIR &  5.39 & $\sim$97.0 & 90.71 & $\sim$87.0 \\\hline
 \citet{polyuhashing} & Iris & VIS\&NIR &  12.41 & $\sim$75.0 & $\sim$64.0 &  $\sim$58.0\\\hline
 \citet{behera2019cross} & Periocular & VIS\&NIR  & 18.79 & 83.12 & x & x \\\hline
 \citet{depresion} & Iris & VIS\&NIR  & \textbf{1.13$\pm$0.14} & \textbf{$\sim$99.9} & $\sim$98.5& $\sim$95.0 \\\hline
 \citet{depresion} & Periocular & VIS\&NIR  & \textbf{0.78$\pm$0.09} & \textbf{$\sim$99.9} & \textbf{$\sim$99.0}&  $\sim$\textbf{98.2} \\\hline

\end{tabular}
\end{adjustbox}
\caption{Summary of Results. *Using only subset of 140 users. $\sim$ values were extracted visually from the graphs in the paper}
\label{table:summary}
\end{table*}

\section{Discussion}
\label{S5: Diss}

In this section, we will comment on the results obtained in the experiments and try to give some insight from them.

First of all, we analyze the performance of the spectrum translation algorithm. From the images presented in Figure \ref{fig:BBDD_perioc} we can appreciate the realistic results obtained. However, some interesting behavior appears in the second fake VIS image, we can see some artifacts in the eyebrow area. These artifacts are most probably because of the reflections at the top of the NIR image, and the interesting part is that the backward conversion can detect and recreate these reflections in the fake image in NIR. In Figure \ref{fig:BBDD_iris}, we can directly see by direct inspection that there are differences from the real and fake images for the iris conversion. There are also some artifacts, vertical lines consequence of the preprocessing steps reflected in \ref{fig:Iris spectrum}, and due to the rearranging of the image in the original shape.

Now we will check the performance based on the results of Table \ref{table:PSNR} and \ref{table:identification}. We analyze the overall visual quality of the transformation in terms of the Peak Signal to Noise Ratio (PSNR) and the Structural Similarity (SSIM). PSNR compares the maximum power that a signal can have and the power of the noise that distorts the signal, while SSIM quantifies the perceptual difference between two images. For the periocular images, we got an average performance of 31.1 dB in PSNR and 0.94 in SSIM for the best conversion. The visual quality assessment is quite stable when comparing each eye. However, we can see an interesting effect, the PSNR is higher for the transformation from VIS spectrum to NIR, but the SSIM metric says that the inverse transform is better. These values are similar to those obtained when applying denoising algorithm to noisy images and compared with the original one e.g., \citet{psnrcnn} \& \citet{refpsnrssim}, although not a fair comparison, it indicates that even with added noise because of the transformation, the overall quality is good. However, the results are not conclusive enough to decide which change would perform better. 

For iris, the quality assessment of the transformation is of special interest, because going from visible to NIR is hard for the GAN because normal limitation of iris texture reflection under visible light condition which can lead to think that the transformation of NIR images to VIS should give a better quality transformation, which agrees with the results. However, the values obtained are not good for whichever transformation and is much worse in comparison with the periocular ones, and a clear indicator that either the data has some issues or the transformation algorithm can't deal with this type of data. Two possible explanations for this can be that either the quality of the data is quite low and this affects the performance of the GAN because it cannot find a good mapping between the spectrum, or this specific GAN architecture cannot reconstruct the high-frequency components characteristic of the iris texture. The first one seems quite plausible, taking into account that the creators of the database dropped 69 subjects out of the 209 because the pictures were of bad quality.

In terms biometric quality, when the CNN was trained with real periocular data and tested with fake data, the grayscale images is the option that gives overall best performance for biometric identification, which is in contrast the option that gave worse results in terms of visual quality. However, when we later on continue with the biometric verification algorithm, for the baseline algorithm, Softmax and Triplet Loss approach, the option that gives the best performance for cross-spectral matching is when using NIR images. For iris recognition, based on the drop of accuracy, the best performing option is when the CNN uses the NIR images, which shows that the spectrum translation of the GAN is actually better to recreate NIR texture for the iris than going from NIR to VIS, which seemed like an easier option. One thing clear along this whole set of experiments is that training the CNNs using fake data improves the overall accuracy, this can be interpreted as applying regularizer to the input of the CNN, although it reduces the performance of the same spectrum data, it improves the performance for cross-spectral. This can also be seen in the later experiments done with Triplet Loss networks in Table \ref{table:verification periocular}.

One behavior that we can appreciate in the verification systems for periocular recognition is that even though in terms of EER, Softmax and Triplet Loss networks give similar results, the latter ones seem to be more stable, providing better performance under more restrictive FAR. One possible explanation of this is the exponential behavior of the Softmax function, and since TL networks use distance metric, it shows a more progressive degradation. However, for iris, the Softmax options constantly gives better results than the Triplet Loss networks for every metric.

Based on the final summary of the results shown in Table \ref{table:summary}, we could improve previously reported results for this dataset except for \citet{depresion}. It is worth to mention that we even achieved that by using  just a fusion of baseline algorithms. However, the performance of just SIFT is still close to the one reported in \citet{polyuhashing} when they used the subset of 140 users and much better than when using the whole database. In comparison with the state of the art results of achieved by \citet{depresion}, where they used pretrained models with a double fine-tuning process for face recognition with more than 13.3M images in total, we trained our CGAN model from scratch, and the CNNs used for biometric recognition were just trained for the image-net dataset and fine-tuned for this database. Still, we achieve a recognition difference of only 0.22\%.



\subsection{Conclusions and Future Work}

In this paper, we investigate the use of Conditional Adversarial Networks for spectrum translation between near infra-red and visual light images with a special focus on ocular biometrics. We analyze the transformation based on the overall visual quality of the transformed images and the accuracy drop of the biometric identification system when trained with opposing data.

We propose two different systems for biometric verification, the first one based on a Siamese Networks trained with Softmax and Cross-Entropy loss, and the second one a Triplet Loss network trained to create embedding of genuine vectors that are close to each other while pushing impostor vectors apart. With the particularity that the CNNs weights used for feature extraction are frozen, and trained for biometric identification for only 1 type of data. We do this to reduce computation time and resources, and we believe that it also helps the network to focus on biometric treats instead of alignment and scale differences, specially for the Triplet Loss networks.

Even though we do not achieve state of the art performance, we could achieve close results using a more reduce training scheme. However, we still outperform most of the previously reported results when we used the periocular area, even in less favorable conditions in some cases, by maintaining the low quality images present in the database or even just using a fusion of baseline algorithms.

For future work, we will try different CGANs for image-to-image translation, with particular interest of any that includes a biometric recognition loss to the optimization of the generator. Training CNNs for identification including real and fake images in the process, and evaluate how the performance change. We will also try alternatives to softmax output for verification due to its less stable performance. Finally, we are also interested in evaluating the systems for verification when the layers for feature extraction are not frozen.

\section{Acknowledgments}

The experimentation of this work was done in a Nvidia Titan V, a donation given by the NVIDIA Corporation.






\bibliographystyle{elsarticle-num-names}
\bibliography{sample.bib}







\end{document}